\documentclass[journal]{IEEEtran}

\usepackage{cite}
\usepackage{amsmath,amssymb,amsfonts}
\usepackage{graphicx}
\usepackage{textcomp}
\usepackage{algorithm}
\usepackage{array}
\usepackage[caption=false,font=normalsize,labelfont=sf,textfont=sf]{subfig}
\usepackage{stfloats}
\usepackage{url}
\usepackage{verbatim}
\usepackage{calc}
\usepackage{oubraces}
\usepackage{algpseudocode}
\usepackage[multiple]{footmisc}
\usepackage{soul}
\usepackage{array,multirow}
\usepackage{booktabs}
\usepackage[normalem]{ulem}
\usepackage{xcolor} 

\def\BibTeX{{\rm B\kern-.05em{\sc i\kern-.025em b}\kern-.08em
    T\kern-.1667em\lower.7ex\hbox{E}\kern-.125emX}}

\begin{document}

\title{Advancing Autonomous Driving: DepthSense with Radar and Spatial Attention}
\author{Muhammad Ishfaq Hussain$^{1}$$^{2}$, Zubia Naz $^{1}$, Muhammad Aasim Rafique$^{3}$, and Moongu Jeon$^{1}$,~\IEEEmembership{Senior Member,~IEEE}%
\thanks{This work was supported GIST - MIT Research Collaboration grant funded by the GIST in  2024.}%
\thanks{$^{1}$ School of Electrical Engineering and Computer Science, Gwangju Institute of Science and Technology, Gwangju, South Korea. e-mail: {\tt\small(ishfaqhussain@gm.gist.ac.kr; zubianaz@gm.gist.ac.kr; mgjeon@gist.ac.kr)}}%
\thanks{$^{2}$ Division of National Science and Technology Data (Large Scale AI Research Group), Korea Institute of Science and Technology Information (KISTI), Daejeon, Rep. of Korea.{\tt\small (ishfaq@kisti.re.kr)}}%
\thanks{$^{3}$ Department of Information Systems, College of Computer Sciences and Information Technology, King Faisal University, Al Ahsa 31982, Saudi Arabia.{\tt\small (mrafique@kfu.edu.sa)}}%
\thanks{ \tt\small{Corresponding Author: mgjeon@gist.ac.kr}}%
}

\maketitle  

\begin{abstract}
Depth perception is crucial for spatial understanding and has traditionally been achieved through stereoscopic imaging. However, the precision of depth estimation using stereoscopic methods depends on the accurate calibration of binocular vision sensors. Monocular cameras, while more accessible, often suffer from reduced accuracy, especially under challenging imaging conditions. Optical sensors, too, face limitations in adverse environments, leading researchers to explore radar technology as a reliable alternative. Although radar provides coarse but accurate signals, its integration with fine-grained monocular camera data remains underexplored. In this research, we propose DepthSense, a novel radar-assisted monocular depth enhancement approach. DepthSense employs an encoder-decoder architecture, a Radar Residual Network, feature fusion with a spatial attention mechanism, and an ordinal regression layer to deliver precise depth estimations. We conducted extensive experiments on the nuScenes dataset to validate the effectiveness of DepthSense. Our methodology not only surpasses existing approaches in quantitative performance but also reduces parameter complexity and inference times. Our findings demonstrate that DepthSense represents a significant advancement over traditional stereo methods, offering a robust and efficient solution for depth estimation in autonomous driving. By leveraging the complementary strengths of radar and monocular camera data, DepthSense sets a new benchmark in the field, paving the way for more reliable and accurate spatial perception systems.
\end{abstract}


\section{Introduction}
 \IEEEPARstart{R}{obots}, the marvels of modern technology, \textcolor{black}{possess} an innate ability to perceive and comprehend their surroundings through an intricate network of sensors. Cameras, radar, lidar, and ultrasonic sensors \textcolor{black}{stand} as the stalwarts in this array, each \textcolor{black}{playing} a pivotal role in capturing vital data and unraveling invaluable insights \cite{b1}, \cite{b25}, \cite{b32}. Among these, cameras \textcolor{black}{reign} supreme, propelled by the strides made in computer vision, which \textcolor{black}{have rendered} them indispensable in granting robots visual acuity through the marvels of AI algorithms \cite{b1}. 
However, despite the wealth of visual information cameras \textcolor{black}{provide}, achieving precise depth estimation \textcolor{black}{remains} essential for tasks like navigation and obstacle avoidance \cite{b1}, \cite{b4}.
\begin{figure}[ht]
    \centering
    \includegraphics[width=70mm]{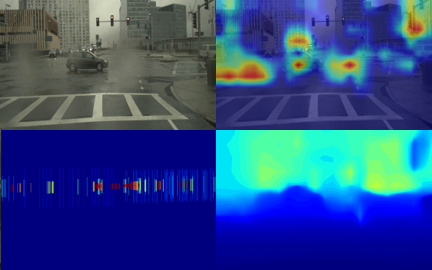}
    \caption{Class activation maps and additional radar markers used in validating depth cues.}
    \label{cam_overview}
\end{figure}
Traditional methods, like stereoscopic vision employing stereo cameras, \textcolor{black}{demand} meticulous calibration and \textcolor{black}{exhibit} limited adaptability to diverse applications \cite{b2},\cite{b5}. Moreover, the high cost associated with lidar often \textcolor{black}{renders} it unattainable for many \cite{b2}. Herein \textcolor{black}{lies} the allure of monocular camera depth estimation, offering a cost-effective alternative \cite{b2}, albeit not without its challenges stemming from the inherent lack of depth cues from a single view \cite{b4}. Depth estimation \textcolor{black}{is} formulated as a per-pixel regression task, further enhanced through ordinal regression techniques to accommodate the sparse nature of radar data \cite{b8}. Various data fusion strategies, including early, mid, and late fusion techniques, \textcolor{black}{are explored}, with empirical evidence favoring late fusion for optimal performance \cite{b12}. Since radar data \textcolor{black}{is} sparse and \textcolor{black}{provides} coarser data than the RGB image data, it \textcolor{black}{is} often fused with RGB using a late fusion technique, where the RGB data \textcolor{black}{is encoded} to a lower modality and \textcolor{black}{concatenated} with sparse radar data \cite{b5}, \cite{b6}. \textcolor{black}{While monocular depth estimation presents a viable solution, it lacks inherent depth cues and struggles in challenging environmental conditions \cite{b9}. Radar, on the other hand, offers precise depth information and serves as a cost-effective and standard sensor in the robotics suite compared to Lidar. Integrating radar data with images through data fusion not only validates visual features but also enhances depth estimation accuracy \cite{b12}.}

The main contributions of this study are:

\begin{enumerate}
    \item A novel deep encoder-decoder network approach is proposed aimed at extracting deep features from RGB images.
    \item The Radar Residual Network is implemented to harness unique features from radar point cloud data. In the quest for an optimal fusion technique, this work explores early, mid, and late data fusion strategies.
    \item A neural fusion technique is introduced, seamlessly integrating radar and image data with precise focus on salient information through a spatial attention mechanism (SAM) to extract in-depth features.
    \item A deep ordinal regression layer is employed, utilizing the spaced-increased discretization technique that enhances feature fusion layers to improve depth completion precision.
    \item Sparse radar data is augmented with fine-grained markers, demonstrating the complementary relationship between radar data and image pyramid features through Class Activation Mapping (CAM). Figure \ref{cam_overview} in the abstract depicts the class activation maps of validated depth cues with additional radar markers.
\end{enumerate}

\section{Related Work}\label{related-work}

This section highlights recent advancements in independent monocular depth estimation and its integration with various data fusion methods. Readers interested in exploring further literature on monocular depth estimation may consult \cite{b28}, \cite{b29}, \cite{b39}, \cite{b40}, {\cite{b42}, \cite{b43}, \cite{b44}, \cite{b45}} and \cite{b41} for in-depth insights.

\subsection{Monocular Depth Estimation}
For robotics and self-driving applications, depth estimation using images offers a cost-effective solution, aligning with advancements in computer vision research. Initially, depth estimation for understanding the 3D world relied on stereo images coupled with deep neural networks \cite{b18}, \cite{b28}, \cite{b29}. However, the limited applicability of stereo images led to a demand for more affordable, robust, and simplified monocular depth estimation methods. Since depth estimation with a single camera poses challenges due to its ill-posed nature and the absence of labeled datasets, early solutions in pattern recognition suggested handcrafted feature extraction for monocular depth estimation \cite{b19}, \cite{b20}. Subsequent advancements in deep learning, coupled with the availability of labeled datasets \cite{b21}, \cite{b22}, spurred the adoption of convolutional neural networks (CNNs) for depth prediction \cite{b23, b24}. {The authors in} \cite{b10} introduced a plug-and-play module that enhances monocular depth prediction by integrating sparse depth inputs into pre-trained models without additional training. A series of CNN-based approaches for depth estimation using monocular images showcased their efficacy in challenging robotics environments \cite{b3}, \cite{b7}, \cite{b25}, \cite{b34}, \cite{b35}, \cite{b36}. Notably, methods leveraging dense representations \cite{b26} through skip connections were able to generate multi-scale feature maps for depth prediction, often employing the common $L1$ loss\cite{b32}. For instance, \cite{b3} introduced a deep neural network with ordinal regression loss, transforming regression into a classification task through spaced-increased discretization (SID). Our approach differs by incorporating radar data into the depth estimation process, which enhances accuracy in low-visibility conditions, as discussed in Section III.
\subsection{Validation using Radars Markers}
Recent works, such as \cite{b9}, \cite{b12}, and \cite{b15}, have leveraged deep learning-based approaches for object detection, incorporating radar's point cloud data as an additional channel alongside monocular images. Their findings inspired our investigation into utilizing radar as an additional sensor for depth estimation in challenging scenarios. In \cite{b2}, various fusion methods of RGB and radar data for depth estimation were explored, employing an encoder-decoder architecture with CNN models to assess early fusion, mid-fusion, late fusion, and multi-level fusion results. Their study suggests that late fusion and multi-layer techniques are better suited to handle the sparsity of radar data. Similarly, \cite{b17} proposed a fusion solution using RGB and radar data, extracting features from both modalities using a region proposal network (RPN) and detection heads, which then generate point cloud data-guided regions of interest (ROIs). For monocular depth estimation with radar data, \cite{b5} adopted a deep ordinal regression network (DORN), considering limited field-of-view scenarios and employing early and late fusion approaches. They augmented radar data with extended data markers through a heuristic method, leading to improved results across diverse weather conditions. While most radar-camera fusion strategies focus on fusion at the detection stage, \cite{b16} pursued a pixel-level fusion approach. They addressed mapping issues with two-stage architectures, associating radar depth with image pixels in the first stage to convert the data into multi-channel enhanced radar (MER) data. This MER data is then fed into another DNN for depth completion in the second stage. Furthermore, another study by \cite{b6} utilized a self-supervised approach for depth estimation tasks, with radar data acting as a weak supervision signal during training. They proposed making radar data optional during inference to mitigate inherent noise and data sparsity issues. The authors in \cite{cvpr2023} propose a method that improves dense depth estimation by mapping each radar point to possible image surfaces and selectively fusing radar and camera data using a gated fusion scheme. Our proposed research builds upon these methods by introducing a novel late fusion technique that enhances depth estimation performance, as detailed in Section III, Part D. We also conducted a comparative study on runtime and memory requirements, which is presented in Section IV, Part B. Additionally, an ablation study was performed to evaluate the impact of sensor failure or the unavailability of radar sensors on depth estimation accuracy, as discussed in Section IV, Part C.

\section{Proposed Methodology}\label{proposed-methodology}
\begin{figure*}[!t]
\centerline{\includegraphics[width=\textwidth]{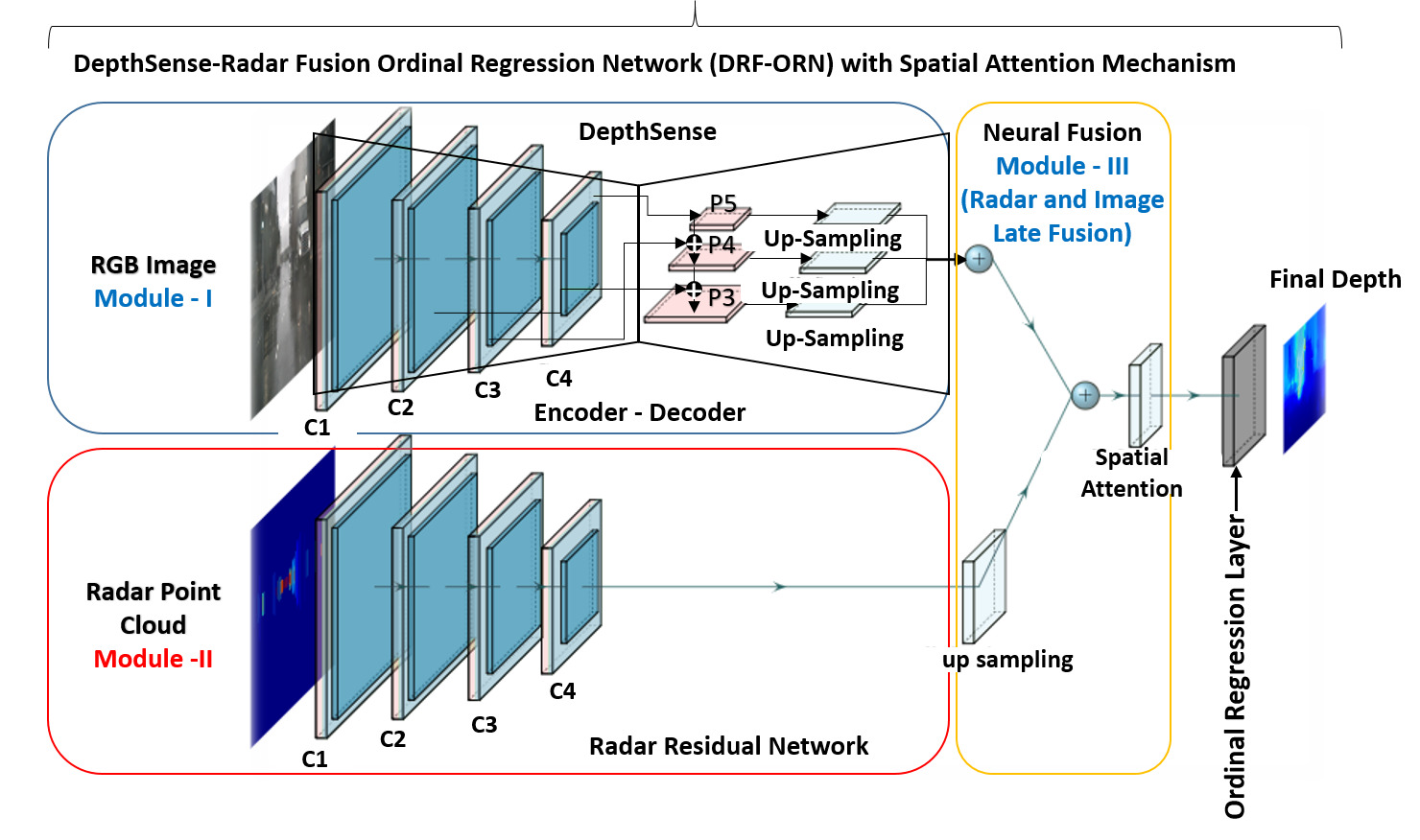}}
\caption{An overview of proposed models structure. Late-fusion technique is applied to extracted features from both the binocular modalities. At the end an ordinal regression layers is applied for Monocular depth estimation.}
\label{structure}
\end{figure*}

\begin{figure}[!h]
\centerline{\includegraphics[width=8.5cm, height=3.5cm]{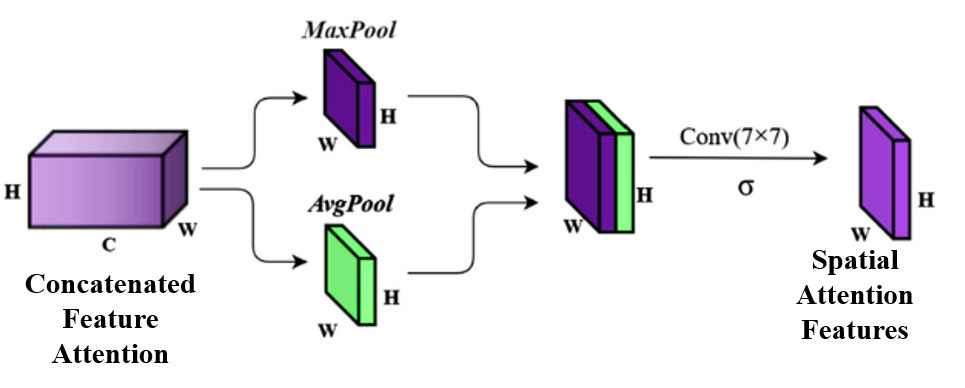}}
\caption{The concatenated features processed through spatial Attention module (SAM) utilized on top of the concatenation process to extract the in depth features.}
\label{structure-attention}
\end{figure}

\textcolor{black}{The proposed methodology consists of four key modules: a deep \textcolor{black}{encoder-decoder} network, a CNN-based \textcolor{black}{radar residual network}, \textcolor{black}{a neural fusion with the spatial attention mechanism}, and an ordinal regression layer, as illustrated in Fig. \ref{structure}. Each module is elaborated upon in the subsequent subsections. The RGB image data is processed through a deep \textcolor{black}{encoder-decoder} network \cite{b13}, with a residual network as its backbone \cite{b11}, while the radar data is processed through a dedicated residual block to extract radar-specific features. The features extracted from both the RGB image and radar data are fused and processed through the spatial attention mechanism, as shown in Fig. \ref{structure-attention}, and then forwarded to the ordinal regression layer based on \cite{b7}. The composition of these modules forms a novel network architecture that aims to enhance performance compared to existing works. By leveraging the DepthSense network for processing RGB images and incorporating radar data through a ResNet-based CNN, our approach combines the strengths of both modalities. The fusion of information {enhances} the capabilities of traditional methods. Additionally, the integration of the ordinal regression layer enables our network to {improve accuracy and robustness in depth prediction tasks}. The subsequent sections provide detailed explanations of each module}.

\subsection{DepthSense Network}\label{rgb-network}
\textcolor{black}{In this study, we depart from traditional Feature Pyramid Networks (FPN) and instead adopt a deep \textcolor{black}{encoder-decoder} network architecture to tackle the challenge of monocular depth estimation. Unlike FPN, our approach encodes input features into a latent representation and decodes them back into the desired output, offering a novel strategy for feature extraction and integration. Leveraging ResNet as the backbone, our network utilizes three pyramid feature layers to capture multi-scale information.  The architecture difference of our study compared to \cite{b14} is hinged on how the pyramid levels are computed using top-down and lateral connections. The pyramid layers from P3 to P5 (refer to Fig. \ref{structure}) are linked to corresponding layers in the residual network (C2 to C4) in reverse order.}

\textcolor{black}{Specifically, P3 is derived from the \textcolor{black}{C2} feature maps with a top-down feed from P5, which in turn receives input from the C4 convolution layer and passes information to P4. Furthermore, P4 processes information from the C3 convolution layer in the encoder, as depicted in Fig. \ref{structure}. The approach implemented in this study is motivated by the fact that the up-sampling on feature pyramids (P3 to P5) accommodates the fine-grained details and reduces the parameters and computation cost of the network compared to the approach described in \cite{b14}.}



\begin{algorithm}
\caption{{DepthSense-Radar Fusion Ordinal Regression Network (DRF-ORN)}}
\begin{algorithmic}[1]

\Require {RGB Image $I_{rgb}$, Radar Point Cloud $I_{radar}$}
\Ensure {Final Depth Map $D$}

\State {\textbf{Step 1: RGB Image Processing (Module I)}}
\State {Extract feature maps from the RGB image through convolutional layers:}
    \State {$C1_{rgb} \gets \text{Conv}(I_{rgb})$}
    \State {$C2_{rgb} \gets \text{Conv}(C1_{rgb})$}
    \State {$C3_{rgb} \gets \text{Conv}(C2_{rgb})$}
    \State {$C4_{rgb} \gets \text{Conv}(C3_{rgb})$}
\State {Perform upsampling on the RGB feature maps:}
    \State {$P3_{rgb} \gets \text{Upsample}(C4_{rgb})$}
    \State {$P4_{rgb} \gets \text{Upsample}(P3_{rgb})$}
    \State {$P5_{rgb} \gets \text{Upsample}(P4_{rgb})$}

\State {\textbf{Step 2: Radar Data Processing (Module II)}}
\State {Extract feature maps from the radar point cloud through convolutional layers:}
    \State {$C1_{radar} \gets \text{Conv}(I_{radar})$}
    \State {$C2_{radar} \gets \text{Conv}(C1_{radar})$}
    \State {$C3_{radar} \gets \text{Conv}(C2_{radar})$}
    \State {$C4_{radar} \gets \text{Conv}(C3_{radar})$}
\State {Perform upsampling on the radar feature maps:}
    \State {$P_{radar} \gets \text{Upsample}(C4_{radar})$}

\State {\textbf{Step 3: Neural Fusion (Module III)}}
\State {Fuse the upsampled RGB and radar features:}
    \State {$F_{fused} \gets \text{Fusion}(P5_{rgb}, P_{radar})$}

\State {\textbf{Step 4: Apply Spatial Attention Mechanism}}
\State {Enhance fused features using spatial attention:}
    \State {$F_{att} \gets \text{SpatialAttention}(F_{fused})$}

\State {\textbf{Step 5: Ordinal Regression for Depth Prediction}}
\State {Predict depth using the ordinal regression layer:}
    \State {$D \gets \text{OrdinalRegression}(F_{att})$}

\State {\textbf{Return:} Final Depth Map $D$}
\end{algorithmic}
\end{algorithm}

\subsection{Radar Residual Network}\label{radar-network}
Given the sparse nature of radar point cloud data, it is standard practice to augment this data to enhance  {its utility and accuracy} \cite{b2}, \cite{b5}. Details of the augmentation process can be found in Section \ref{ssecRda}. In our study, radar data is processed using a residual {standalone} network specifically designed to efficiently extract radar-specific features. This network employs a residual architecture, which is vital in avoiding the masking of coarse radar features by fine-grained RGB data when both are learned together within a CNN framework. To achieve this, {ResNet-18 is utilized} for radar feature extraction. ResNet-18, known for its efficiency and robust feature extraction, employs multiple residual blocks to capture both low- and high-level radar features. By maintaining a separate extraction pathway, the network ensures accurate isolation of radar features, leading to improved performance in subsequent analysis tasks. This approach significantly enhances the precision and reliability of radar feature extraction.

\subsection{Radar and RGB Neural Fusion}

In this study, we adopt a late fusion (neural fusion) technique where radar data is encoded separately using a dedicated residual network \textcolor{black}{(section - \ref{radar-network}), and the camera data is processed through the DepthSense network (section - \ref{rgb-network})}.

The fusion of features from both sensors facilitates the validation of accurate distance information, thereby reinforcing the purposeful depth relationship among RGB features. The fused features are processed through the concatenated feature attention network. The spatial attention mechanism (SAM) \textcolor{black}{\cite{b31}} is utilized to process those radar camera fused features to precisely attain the in-depth features. The SAM contribution in this research further improves the quality of the work. The structure of SAM can be seen in Fig. \ref{structure-attention}.


\begin{figure*}[t]
  \centering
  \includegraphics[width=185mm, height=50mm, scale=0.1]{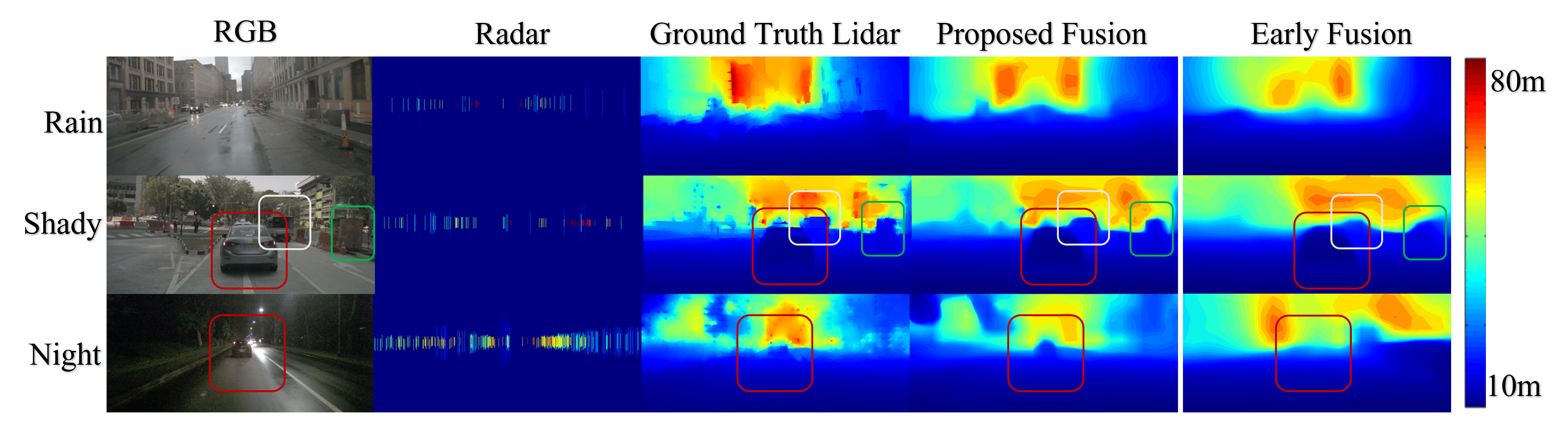}
  \caption{
  The qualitative results of depth estimation using RGB+Radar with only extended radar's point cloud.}
  \label{height-extended}
\end{figure*}
\begin{figure*}[t]
  \centering
  \includegraphics[width=160mm, height=48mm, scale=0.1]{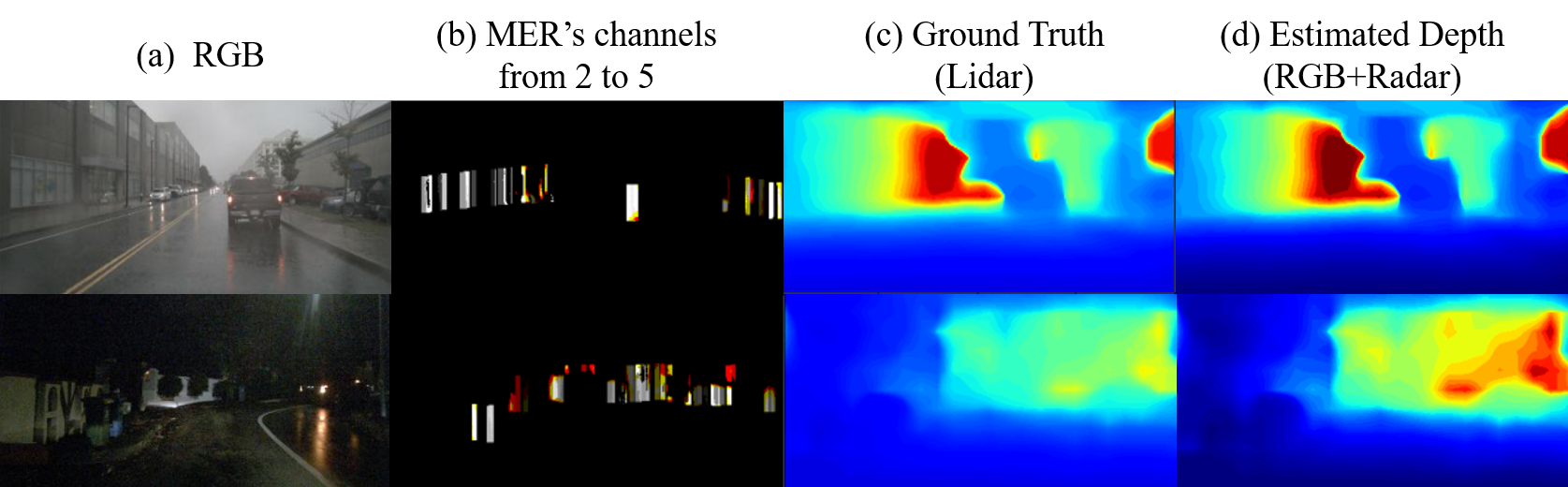}
  \caption{
  The qualitative results are based on RGB+MERs radar point cloud.}
  \label{mers-fig}
\end{figure*}

\subsection{Ordinal Regression Layer}

In monocular depth estimation, ordinal regression resolves challenges stemming from the disparity between predicted and estimated depths in 2D images, which are influenced by perspective projection aspect ratios. By transforming the regression problem into a classification task, ordinal regression divides real-valued distances into bins, facilitating normalized error computation. Adopting an ordinal regression layer composition and loss function as discussed in \cite{b7}, \cite{b5}, this study enhances depth estimation accuracy. Furthermore, the incorporation of a modified feature pyramid network introduces a robust inductive bias into monocular image features, improving depth estimation. Conditioning with radar data further enhances prediction accuracy, particularly in scenarios where image data alone is insufficient. This unified approach represents a significant advancement in \textcolor{black}{state of the art (SOTA)} depth estimation techniques, leveraging simple modules.

\subsection{Radar data augmentation using radar channel Enhancement (MER's)}
\label{ssecRda}

This study delves into extending sparse point cloud radar data for more meaningful interpretation. Initially, we explore height extension of markers, as depicted in the second column of Fig. \ref{height-extended}. A second method involves extending markers with assistance from neighboring pixels in the image, establishing pixels-to-depth association (PDA) using a DNN supervised with lidar point cloud data \cite{b16}. This method expands both the width and height of markers and introduces a separate channel for various confidence levels of association, creating a multiple channel-based enhanced radar image (MER), as illustrated in the \textcolor{black}{second} column of Fig. \ref{height-extended}. MER provides estimates of extended heights and widths of regions potentially occupied by objects in the real scene. While simple height extension elongates radar data in strip form, MER employs a trained ANN to predict extension in neighboring regions, considering confidence levels and thresholds. This comprehensive approach significantly enhances radar data interpretation, paving the way for more accurate depth estimation in complex environments.


\begin{table*}[t]
\caption{These evaluations covered various scenarios categorized into clear day, rain, and night scenarios, all based on the test dataset (compared only with SOTA). The evaluation method employed late fusion, combining both sensor modalities, which include monocular images and radar data.}
\begin{center}
\begin{tabular}{lcccccccccccccc}

\multicolumn{15}{c} {\textbf{Results on Full nuScenes Dataset}} \\
\hline \textbf{Models} & \multicolumn{4}{c} {$\delta_{1}\uparrow$} && \multicolumn{4}{c} {$RMSE \downarrow$} && \multicolumn{4}{c} {$AbsRel \downarrow$} \\
\cline { 2 - 15 } & combine & day & night & rain & & combine & day & night & rain & & combine & day & night & rain \\
\hline RGB only & $0.88$ & $0.88$ & $0.77$ & $0.88$ & & $5.15$ & $5.10$ & $6.79$ & $5.39$ & & $0.11$ & $0.11$ & $0.16$ & $0.10$ \\
Lin et al.- IROS (2020) \cite{b2} & $0.88$ & $0.89$ & $\mathbf{0.81}$ & $-$ & & $5.40$ & $5.27$ & $6.40$ & $-$ & & $0.11$ & $0.10$ & $\mathbf{0.14}$ & $-$ \\
DORN - ICIP (2021)  \cite{b5} & $0.88$ & $0.90$ & $0.78$ & $0.88$ & & $5.19$ & $4.97$ & $6.86$ & $5.48$ & & $0.10$ & $0.10$ & $0.15$ & $0.10$ \\
Stefano, et al. \cite{b6} & $-$ & $0.86$ & $-$ & $-$ & & $-$ & $6.43$ & $-$ & $-$ & & $0.13$ & $0.12$ & $0.21$ & $0.14$ \\
RadarNet,- CVPR (2023) \cite{cvpr2023} & $-$ & $-$ & $-$ & $-$ & & $-$ & $4.898$ & $-$ & $-$ & & $-$ & $-$ & $-$ & $-$ \\
Early Fusion (DepthSense) & ${0.85}$ & ${0.86}$ & ${0.77}$ & ${0.86}$ & & ${6.26}$ & ${6.04}$ & $7.23$ & ${6.21}$ & & ${0.13}$ & ${0.13}$ & ${0.17}$ & ${0.12}$\\
Proposed Fusion (DepthSense) & $\mathbf{0.90}$ & $\mathbf{0.91}$ & $\mathbf{0.81}$ & $\mathbf{0.91}$ & & $\mathbf{4.91}$ & $\mathbf{4.83}$ & $6.57$ & $\mathbf{4.98}$ & & $\mathbf{0.10}$ & $\mathbf{0.10}$ & $\mathbf{0.14}$ & $\mathbf{0.09}$\\
Proposed DepthSense with MERs &  $\mathbf{0.90}$ & $\mathbf{0.91}$ & $0.79$ & $\mathbf{0.91}$ & & $\mathbf{3.54}$ & $\mathbf{3.39}$ & $\mathbf{5.16}$ & $\mathbf{3.40}$ & & $\mathbf{0.10}$& $\mathbf{0.09}$ & $0.15$ & $0.10$\\
\hline
\end{tabular}

\label{height-result}
\end{center}
\end{table*}

\begin{table*}[t]
\caption{Proposed models (DepthSense), results based on Multiple channel based enhanced radar (MER), image.}
\begin{center}
\begin{tabular}{p{2.1cm}p{0.7cm}p{0.3cm}p{0.3cm}p{0.3cm}p{0.4cm}p{0.7cm}p{0.3cm}p{0.3cm}p{0.3cm}p{0.40cm}p{0.7cm}p{0.3cm}p{0.3cm}p{0.3cm}p{0.45cm}p{0.7cm}p{0.3cm}p{0.3cm}p{0.40cm}}
\multicolumn{19}{c} {\textbf{Evaluation on Full nuScenes Dataset with MER's - Estimation error with low height region (0.3 to 2(meters) above the ground level)}} \\
\hline Models & \multicolumn{4}{c} {$\delta_{1}\uparrow$} && \multicolumn{4}{c} {$RMSE \downarrow$} && \multicolumn{4}{c} {$AbsRel \downarrow$} && \multicolumn{4}{c} {$RMSE_{log}\downarrow$}\\
\cline { 2 - 20 } & combine & day & night & rain && combine & day & night & rain && combine & day & night & rain && combine & day & night & rain\\
\hline \textcolor{black}{DORN \cite{b5}}  & $0.74$ & $0.75$ & $0.65$ & $0.77$ && $5.10$ & $5.13$ & $6.42$ & $4.80$ && $0.18$ & $0.18$ & $0.19$ & $0.19$ && $0.24$ & $0.24$ & $0.30$ & $0.24$\\
HourglassNet \cite{mers} - CVPR2021  & $0.81$ & $0.82$ & $0.72$ & $0.80$ && $4.22$ & $4.18$ & $4.87$ & $4.11$ && $0.14$ & $0.14$ & $0.17$ & $0.16$ && $0.20$ & $0.20$ & $0.23$ & $0.21$\\
DepthSense (proposed) & $\mathbf{0.84}$ & $\mathbf{0.84}$ & $\mathbf{0.75}$ & $\mathbf{0.86}$ && $\mathbf{4.15}$ & $\mathbf{4.14}$ & $\mathbf{4.75}$ & $\mathbf{3.95}$ && $\mathbf{0.14}$& $\mathbf{0.13}$ & $\mathbf{0.17}$ & $\mathbf{0.13}$ &&  $\mathbf{0.19}$& $\mathbf{0.19}$ & $\mathbf{0.23}$ & $\mathbf{0.13}$\\
\hline \\ \multicolumn{19}{c} {\textbf{MER's (Multi-Enhanced Radar Channels) - Evaluation on Full-image depth completion errors (m)}} \\
\hline \textcolor{black}{DORN \cite{b5}}  & $0.84$ & $0.85$ & $0.68$ & $0.86$ && $4.44$ & $4.27$ & $5.67$ & $4.23$ && $0.13$ & $0.13$ & $0.18$ & $0.13$ && $0.19$ & $0.19$ & $0.27$ & $0.24$\\
HourglassNet \cite{mers} - CVPR2021 & $0.86$ & $0.87$ & $0.75$ & $0.87$ && $3.98$ & $3.84$ & $5.55$ & $3.86$ && $0.11$ & $0.10$ & $0.16$ & $0.11$ && $0.18$ & $0.17$ & $0.24$ & $0.18$\\
DepthSense (proposed) & $\mathbf{0.90}$ & $\mathbf{0.91}$ & $\mathbf{0.79}$ & $\mathbf{0.91}$ && $\mathbf{3.54}$ & $\mathbf{3.39}$ & $\mathbf{5.16}$ & $\mathbf{3.40}$ && $\mathbf{0.10}$& $\mathbf{0.09}$ & $\mathbf{0.15}$ & $\mathbf{0.10}$ &&  $\mathbf{0.15}$& $\mathbf{0.15}$ & $\mathbf{0.23}$ & $\mathbf{0.15}$\\
\hline
\end{tabular}

\label{MER-Result}
\end{center}
\end{table*}

\begin{table*}[ht]
    \centering
    \caption{Performance Comparison in terms of the Scenario in the similar conditions}
    \begin{tabular}{cccccccc}
        \toprule
        \textbf{Models} & \textbf{Radar Format} & $\delta_1 \uparrow$ & $\delta_2 \uparrow$ & $\delta_3 \uparrow$ & \textbf{RMSE} $\downarrow$ & \textbf{MAE} $\downarrow$ & \textbf{AbsRel} $\downarrow$ \\
        \midrule
        PnP \cite{b10} & None & 0.863 & 0.948 & 0.976 & 5.578 & - & 0.128 \\
        Sparse-to-dense \cite{b3} & None & 0.862 & 0.948 & 0.976 & 5.613 & - & 0.126 \\
        DORN \cite{b5} & None & 0.872 & 0.952 & 0.978 & 5.382 & - & 0.117 \\
        CSPN \cite{b27} & None & 0.882 & 0.958 & 0.985 & 5.385 & - & 0.123 \\
        RCDformer \cite{b19} & None & 0.890 & 0.957 & 0.978 & 5.205 & 2.317 & 0.114 \\
        DepthSense & None & 0.880 & 0.965 & 0.962 & 5.152 & 2.192 & 0.116 \\

        \midrule
        PnP \cite{b10} & Raw & 0.863 & 0.948 & 0.976 & 5.578 & 2.496 & 0.128 \\
        Sparse-to-dense \cite{b3} & Raw & 0.876 & 0.949 & 0.974 & 5.628 & 2.374 & 0.115 \\
        CSPN \cite{b27} & Raw & 0.882 & 0.958 & 0.985 & 5.561 & 2.457 & 0.117 \\
        RadarNet (single stage) \cite{cvpr2023} & Raw & 0.884 & 0.953 & 0.977 & 5.409 & 2.27 & 0.109 \\
        RadarNet (two stage) \cite{cvpr2023} & Raw & 0.889 & 0.961 & 0.984 & 5.180 & 2.061 & 0.101 \\
        Dorn\_radar (single stage) \cite{b5} & Height-extend & 0.889 & 0.961 & 0.984 & 5.191 & - & 0.100 \\
        Dorn\_radar (two stage) \cite{b5} & Height-extend & 0.895 & 0.958 & 0.978 & 5.206 & - & 0.104 \\
        Lin\_et-al \cite{b2} & Height-extend & 0.880 & 0.950 & 0.970 & 5.270 & - & 0.100 \\        
        Lee et al. \cite{b39} & Raw & 0.895 & 0.958 & 0.978 & 5.209 & 2.104 & 0.100 \\
        RDNet \cite{b41} & Raw & 0.897 & 0.960 & 0.980 & 5.180 & - & 0.100 \\
        RC-PDA \cite{mers} & MER & 0.830 & 0.917 & 0.956 & 6.943 & - & 0.173 \\
        RCDPT \cite{b29} & MER & 0.901 & 0.961 & 0.981 & 5.165 & - & 0.095 \\
        RadarNet. \cite{cvpr2023} & Quasi-Dense Depth & - & - & - & 4.898 & 2.179 & - \\
        RCDformer \cite{b19}  & Height-extend & 0.907 & 0.965 & 0.981 & 5.014 & 2.117 & 0.101 \\
        RCDformer \cite{b19}  & MER & 0.909 & 0.972 & 0.989 & 4.912 & 2.033 & 0.093 \\
        \midrule
        DepthSense (Proposed) & Height-extend & 0.910 & 0.965 & 0.981 & 4.830 & 2.066 & 0.100 \\
        DepthSense (Proposed) & MER & \textbf{0.910} & \textbf{0.980} & \textbf{0.992} & \textbf{3.391} & \textbf{1.989} & \textbf{0.090} \\
        \bottomrule
    \end{tabular}
    \label{tab:performance}
\end{table*}

\begin{table*}[ht]
    \centering
    \caption{ Performance Comparison with respect to different depth in meters}
    \begin{tabular}{cccccc}
        \toprule
        \textbf{Max Eval Distance} & \textbf{Models} & \textbf{\# Radar frames} & \textbf{\# Images} & \textbf{MAE $\downarrow$} & \textbf{RMSE $\downarrow$} \\
        \midrule
        \multirow{5}{*}{50m} & RC-PDA \cite{mers} & 5 & 3 & 2225.0 & 4156.5 \\
        & RC-PDA with HG \cite{mers} & 5 & 3 & 2315.7 & 4321.6 \\
        & DORN \cite{b5} & 5$\times$3 & 1 & 1926.6 & 4124.8 \\
        & RadarNet \cite{cvpr2023} & 1 & 1 & \textbf1727.7 & 3746.8 \\
        & DepthSense (Height extended) & 1 & 1 & 1644.5 & 3609.2 \\
        & DepthSense (MERs) & 5 & 1 & \textbf{1615.4} & \textbf{2940.1} \\
    
        \midrule
        \multirow{5}{*}{70m} & RC-PDA \cite{mers} & 5 & 3 & 3326.1 & 6700.6 \\
        & RC-PDA with HG \cite{mers} & 5 & 3 & 3485.6 & 7020.9 \\
        & DORN \cite{b5} & 5$\times$3 & 1 & 2380.6 & 5252.7 \\
        & RadarNet \cite{cvpr2023} & 1 & 1 & 2073.2 & 4590.7 \\
        & DepthSense (Height extended) & 1 & 1 & 1939.7 & 4495.0 \\
        & DepthSense (MERs) & 5 & 1 & \textbf{1829.6} & \textbf{3112.2} \\
        \midrule
        \multirow{9}{*}{80m} & RC-PDA \cite{mers} & 5 & 3 & 3713.6 & 7692.8 \\
        & RC-PDA with HG \cite{mers} & 5 & 3 & 3884.3 & 8008.6 \\
        & DORN \cite{b5} & 5$\times$3 & 1 & 2467.7 & 5554.9 \\
        & Lin \cite{b2} & 3 & 3 & 2371.0 & 5623.0 \\
        & R4Dyn \cite{b6} & 4 & 3 & N/A & 6434.0 \\
        & Sparse-to-dense \cite{b3} & 5 & 1 & 2374.0 & 5628.0 \\
        & PnP \cite{b10} & 4 & 1 & 2496.0 & 5578.0 \\
        & RadarNet \cite{cvpr2023} & 1 & 1 & 2179.3 & 4898.7 \\
        & DepthSense (Proposed) (Height extended) & 1 & 1 & 2066.1 & 4910.4 \\
        & DepthSense (Proposed) with (MERs) & 5 & 1 & \textbf{1989.5} & \textbf{3391.8} \\

        \bottomrule
    \end{tabular}
    \label{tab:performance:in:meters}
\end{table*}

\section{Experimental Work}\label{experimental-work}

\subsection{Dataset}
All experiments in this study were conducted using the nuScenes dataset \cite{b32}, chosen for its comprehensive coverage of various environmental conditions. Spanning nearly 15 hours of driving across Singapore and Boston roads, the dataset encompasses diverse scenarios such as rain, night, daylight, and varying illuminations. Each scene captures data from 1 lidar, 6 cameras, and 5 radars, providing a 360-degree field of view. With 1000 scenes, of which 850 are annotated, the dataset offers ample data for robust experimentation. An average of 40 samples are drawn every 20 seconds of driving from each scene. Furthermore, the nuScenes dataset includes a development kit that facilitates sensor calibration and data projection. Similar to prior works such as \cite{b2}, we utilized 850 labeled scenes, dividing them into 750 training, 15 validation, and 85 test sets for fair comparison with the SOTA. Training involved 30,736 sample frames from the selected scenes, with 3,424 sample frames reserved for evaluation. In this study, we utilized front camera images and overlapped radar point cloud data. Notably, the official dataset lacks depth annotations; hence, we followed the approach in \cite{b2} and utilized lidar point cloud information as ground truth \cite{b37}. This dataset's extensive and diverse nature, coupled with its availability of large-scale annotated data, makes it an ideal choice for our depth estimation experiments, enabling rigorous evaluation and comparison against existing methodologies.

\subsection{Implementation Details}

The experiments were conducted on a single machine equipped with a 24GB Nvidia-GTX-3090 GPU and 48GB of internal memory, leveraging the PyTorch framework \cite{b33}. During model training, a batch size of 8 was utilized, with stochastic gradient descent (SGD) serving as the optimizer. Initially, the learning rate was set to 0.001 with a polynomial power of 0.9, while the momentum was maintained at 0.9, accompanied by a weight decay of 0.0001 after every ten epochs. The backbone network was initialized with ImageNet pre-trained weights for RGB data, while random weights were assigned to sparse radar inputs. Training extended over 40 epochs, employing ordinal regression as space increasing discretization (SID), following the approach outlined in \cite{b7}, utilizing an ordinal regression loss function \cite{b7}. Consistent with prior work \cite{b7}, 80 intervals were used for the ordinal regression, balancing quantization error and non-discretization. To ensure fair comparison with SOTA techniques, identical parameters were employed across all experiments. Training and evaluation were conducted on down-scaled images from 900 x 1600 to 450 x 900 resolution, with an additional truncation of the top 100 pixels in height. This resolution was maintained for projecting radar points and generating ground truth data using lidar. These experimental protocols were meticulously designed to facilitate reproducibility and enable meaningful comparison with existing methodologies, contributing valuable insights to the depth estimation research community.

\subsection{Experimentation and Results}
\label{ssecExp}
In this study, we compare our proposed monocular depth estimation approach using radar data to \textcolor{black}{state-of-the-art (SOTA)} methodologies. To ensure a robust evaluation, our experiments encompass all scene categories within the dataset, including day, night, and rain scenarios \cite{b32}. These diverse scenarios enable a comprehensive assessment of the effectiveness of our proposed strategy across varying environmental conditions.

\begin{table}[t]
\caption{Results on Full nuScenes Dataset with \textbf{MER's - Evaluation on Low-height region (0.3 to 2(meters) above the ground level)}.}
\begin{center}
\begin{tabular}{|c|c|c|c|c}
\multicolumn{4}{c}{}\\
\hline \textbf{Models} & \textbf{$\delta_{1}<1.25$} & $ \delta_{2}<1.25^2$ & $\delta_{3}< 1.25^3$ \\
\hline \textcolor{black}{Cho et al. \cite{b5}} with MER & $0.74$ & $0.91$ & $0.96$\\
\hline Hourglass \cite{b30} with MER & $0.81$ & $0.94$ & $0.97$\\
\hline (DepthSense) with MER & $\mathbf{0.84}$ & $\mathbf{0.95}$ & $\mathbf{0.98}$ \\
\hline
\end{tabular}
\end{center}
\label{delta-result}
\end{table}

\begin{table}[t]
\caption{On Full nuScenes Dataset with \textbf{MER's - Evaluation on Full-image depth completion errors (m)}}
\begin{center}
\begin{tabular}{|c|c|c|c|c}
\multicolumn{4}{c}{}\\
\hline \textbf{Models} & \textbf{$\delta_{1}<1.25$} & $ \delta_{2}<1.25^2$ & $\delta_{3}< 1.25^3$  \\
\hline \textcolor{black}{Cho et al. \cite{b5}} with MER & $0.84$ & $0.94$ & $0.97$\\
\hline Hourglass \cite{b30} with MER & $0.86$ & $0.95$ & $0.98$\\
\hline (DepthSense) with MER & $\mathbf{0.91}$ & $\mathbf{0.97}$ & $\mathbf{0.99}$ \\
\hline
\end{tabular}
\end{center}

\label{delta-result-mers}
\end{table}
Our experiments are categorized into two main groups: the first group utilizes height-extended radar markers, while the second group employs multiple-channel enhanced radar (MERs), as outlined in Section \ref{proposed-methodology}. Both experiments utilize the proposed architecture with identical hyperparameters, varying solely by the input channel count in the first convolutional layer of the backbone CNN to support MER data input. Notably, our proposed architecture comprises 68 million trainable parameters, approximately half that of a comparable architecture in \cite{b5}. In this study, MERs compute six confidence intervals, with values sourced from \cite{b16}. Furthermore, our approach demonstrates impressive efficiency, with an inference time of 0.118 seconds for a batch size of three on the utilized GPU, significantly outperforming existing models, such as \cite{b5}, which require 0.221 seconds. These findings not only validate the effectiveness of our proposed approach but also highlight its potential for practical implementation, offering substantial gains in both performance and efficiency for real-world applications.

A detailed comparison with state-of-the-art (SOTA) monocular depth estimation techniques, incorporating radar data fusion across various environmental conditions, is presented in Table \ref{height-result}. Evaluation metrics include Root Mean Square Error (RMSE) in equation (\ref{eqn1}), RMSE logarithm (RMSE$_{log}$), and Absolute Relative Error (AbsRel) in equation (\ref{eqn2}), and threshold values $\delta{1}$, $\delta_{2}$, and $\delta_{3}$ \cite{b2} in equation (\ref{eqn3}). $N$ represents the total number of pixels in each depth image, $y$ stands for the ground truth, $\tilde{y}$ denotes the predicted depth pixels, and $n$ is used for the threshold limits in equations (\ref{eqn1}), (\ref{eqn2}), and (\ref{eqn3}).

Quantitative results obtained using height-extended radar markers as input are summarized in Table \ref{height-result}, while Table \ref{MER-Result} {presents results for depth estimation with MER inputs}. The performance of compared techniques is assessed using the provided code from the respective works. Only Chen et al. \cite{b5} provides experimental results on the complete nuScenes dataset, while Juan et al. \cite{b2} focuses exclusively on day and night scene results. Our proposed model consistently outperforms existing techniques across various scenarios, significantly improving performance metrics and reducing inference time. Threshold-based evaluation results ($\delta$) are detailed in Tables \ref{delta-result}, \ref{tab:performance}, \ref{tab:performance:in:meters} and \ref{delta-result-mers}, showcasing quantitative improvements across all evaluation metrics. Additionally, qualitative results are illustrated in Fig. \ref{height-extended} and Fig. \ref{mers-fig}. The advantage of our proposed method over the Hourglass network \cite{b30} is likely due to the simplicity of our model and its suitability for the target problem. While further experimentation in different domains may provide additional insights, we acknowledge the relevance of the no-free-lunch theorem in this context, emphasizing the need for careful consideration of model complexity and problem-specific requirements in future research endeavors.

\textcolor{black}{Our experiments demonstrate that our proposed architecture delivers superior performance with significantly fewer parameters (compared to the architecture in \cite{b5}) and a faster inference time of 0.118 seconds per batch, nearly halving the processing time required by existing models. The efficiency of our model is due to its streamlined design and effective use of a spatial attention mechanism for data fusion. By optimizing the number of input channels and simplifying the backbone CNN architecture, our model not only reduces computational load but also enhances depth estimation accuracy across various environmental conditions. These results highlight the architectural advantages of our approach, offering substantial improvements in both speed and accuracy for real-world applications, and demonstrate how the use of spatial attention mechanisms for fusion contributes to its superior performance compared to existing methods.}


\begin{equation}
\label{eqn1}
\text { Threshold: }  \delta_{n}\\
\qquad \delta_{n}=\left|\left\{\tilde{y} :\max\left(\frac{\tilde{y} }{y },\frac{y }{\tilde{y} }\right)<1.25^{n}\right\}\right| /N   
\end{equation}

\begin{equation}
\label{eqn2}
\text {RMSE:} \\
\qquad \sqrt{\frac{1}{N} \sum\left\|y -\tilde{y} \right\|_{2}^{2}} 
\end{equation}

\begin{equation}
\label{eqn3}
\text {AbsRel:} \\
\qquad \begin{array}{l}
N \sum|y-\tilde{y}| / y .
\end{array}  
\end{equation}




\section{Ablation Study}\label{ablation-study}

Two ablation studies were conducted to further elucidate the efficacy of our proposed method. Firstly, the model's generalization capability was assessed by training it on scenes from various imaging conditions within the nuScenes dataset and testing it on scenes with different environmental characteristics. Table \ref{Reduction-training-data-full} presents qualitative results obtained by reducing the training data of rain, day, and night scenes. Despite training with reduced data, the proposed model which uses MERs input demonstrated consistent performance across scenes, showcasing its robust generalization ability.

\begin{figure*}[!ht]
  \centering
\subfloat[\footnotesize Camera image]{\includegraphics[width=44mm , height=20mm, scale=0.15578]{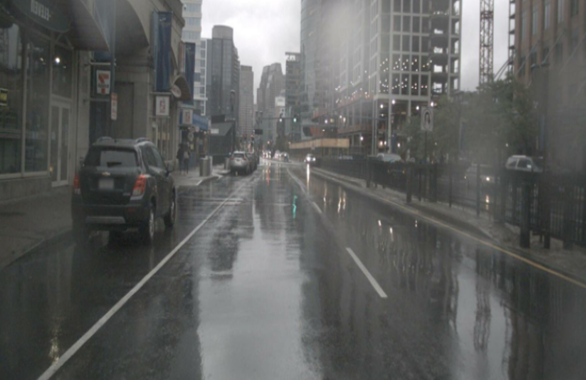}}\subfloat[\footnotesize RaDorn Distance at 33m]{\includegraphics[width=44mm , height=20mm, scale=0.15578]{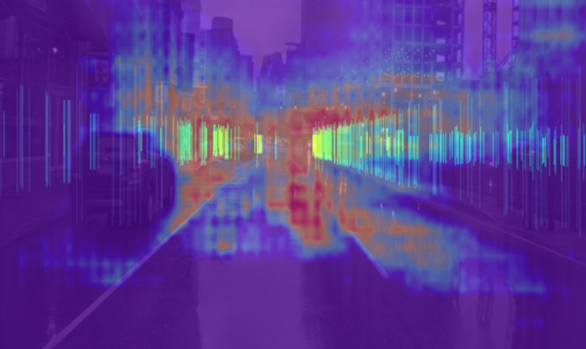}}\subfloat[\footnotesize RaDorn Distance at 55m]{\includegraphics[width=44mm , height=20mm, scale=0.15578]{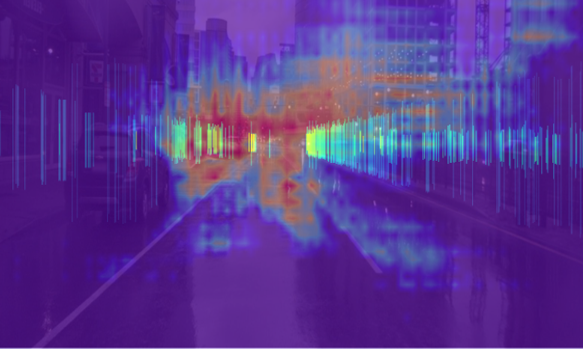}}\subfloat[\footnotesize RaDorn Distance at 70m]{\includegraphics[width=44mm , height=20mm, scale=0.15578]{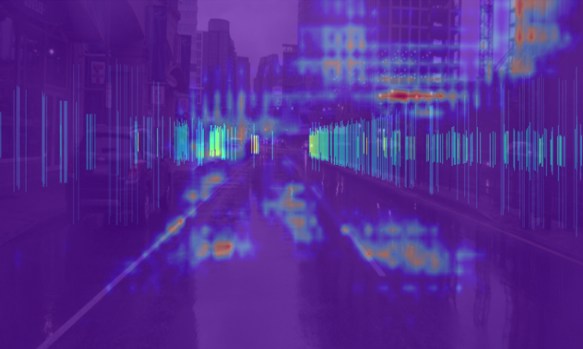}}

\subfloat[\footnotesize Height extended Radar Markers]{\includegraphics[width=44mm , height=20mm, scale=0.15578]{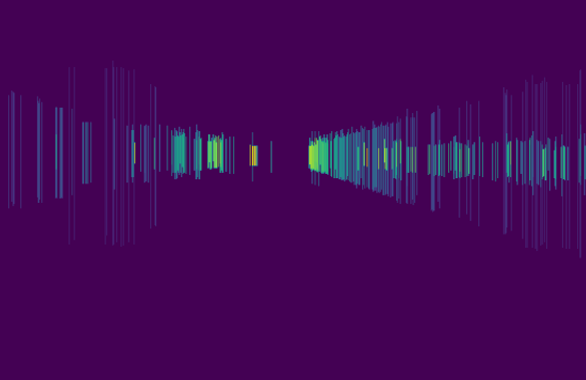}}\subfloat[\footnotesize DepthSense Distance at 35m]{\includegraphics[width=44mm , height=20mm, scale=0.15578]{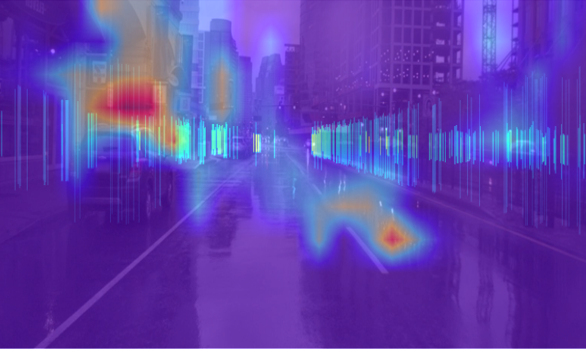}}\subfloat[\footnotesize DepthSense Distance at 55m]{\includegraphics[width=44mm , height=20mm, scale=0.15578]{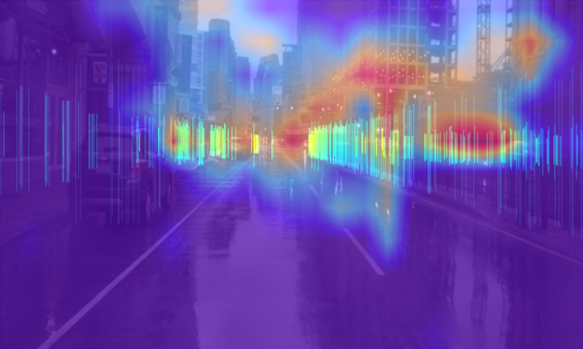}} \subfloat[\footnotesize DepthSense Distance at 70m]{\includegraphics[width=44mm , height=20mm, scale=0.15578]{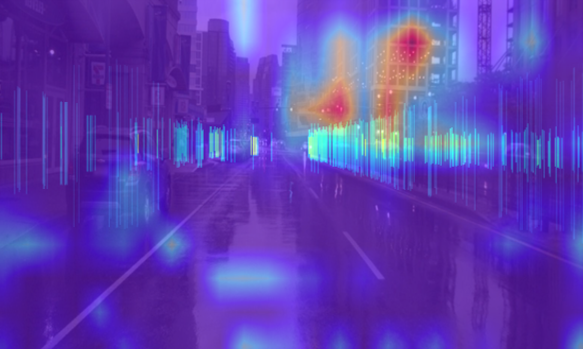}} \caption{\textcolor{black}{The radar markers are projected to visualise the effect of radar information at different distance by using CAM approach. (a) is RGB test image and (b), (c), (d) are RaDorn results with distance values of 35m, 55m, and 70m respectively. (e) shows calibrated height extended radar points for corresponding test image. (f), (g) and (h) depicts activation maps generated with results computed with the proposed DepthSense at same distance values.}}
  \label{cam_comparison}
\end{figure*}

In the second experiment, the loss function was altered to assess the impact on depth estimation accuracy. Specifically, L1 (Mean Absolute Error)  and L2 (Mean Square Error) regression losses were employed in place of ordinal regression loss. However, the model failed to produce equivalent results, underscoring the importance of ordinal regression loss in our proposed technique. Table \ref{change_of_loss} provides qualitative results obtained with alternate regression loss functions, reinforcing the superiority of ordinal regression loss for accurate depth estimation. These ablation studies provide valuable insights into the critical components of our proposed method and underscore its effectiveness in addressing depth estimation challenges.
\subsection{Class Activation Maps}
In addition, we conducted an analysis to evaluate the impact of incorporating radar markers and pyramid feature layers using activation maps. Leveraging ordinal regression, we employed GRAD-CAM++ \cite{cam} with distance classes to visualize the influence of radar data on depth estimation. GRAD-CAM++ utilizes derivatives of convolution feature maps weighted by scores of specific distance classes to pinpoint regions contributing to the model's decisions. Fig. \ref{cam_comparison} illustrates CAM results for monocular depth estimation with and without radar data, focusing on frames with distance values of 35m, 55m, and 70m. Our findings reveal that depth estimation with our proposed method benefits significantly from radar markers, which actively contribute to depth estimation by validating regions crucial for accurate depth perception. Conversely, depth estimation without radar data is influenced primarily by the vanishing point phenomenon in perspective projection. Furthermore, the pyramid feature network preserves spatial locality and scale invariance properties, enhancing the fusion of sensor inputs. These CAM tests offer valuable insights into the efficacy of individual sensors' inputs in data fusion, particularly highlighting the effectiveness of radar markers in addressing challenges associated with ill-posed monocular depth estimation problems.

\begin{table}[ht]
\caption{\textcolor{black}{Reducing training nuScenes Dataset (Day, Night, Rain) separately with \textbf{MER's - Evaluation on Full-image depth completion errors (m)}}}
\begin{center}
\begin{tabular}{|c|c|c|c|c| c}
\multicolumn{5}{c}{{\textbf{Different training and testing scenes experiment for ablation study }}}\\

\hline \textbf{Methods} & \textbf{$\delta_{1}\uparrow$} & {$RMSE \downarrow$} & {$AbsRel \downarrow$} & {$RMSE_{log}\downarrow$}  \\
\hline Day & $0.88$ & $3.76$ & $0.11$ & $0.17 $\\
\hline Night& $0.70$ & $5.73$ & $0.19$ & $ 0.25$\\
\hline Rain& $0.87$ & $3.96$ & $0.12$ & $0.18 $\\
\hline
\end{tabular}
\end{center}

\label{Reduction-training-data-full}
\end{table}

\begin{table}[ht]
\caption{\textcolor{black}{Applying Different Loss Function}}
\begin{center}
\begin{tabular}{|c|c|c|c|c| c}
\multicolumn{5}{c}{\textbf{Proposed Model (DepthSense) on Full nuScenes with different Loss}}\\ 

\hline \textbf{Loss Function} & \textbf{$\delta_{1}\uparrow$} & {$RMSE \downarrow$} & {$AbsRel \downarrow$} & {$RMSE_{log}\downarrow$}  \\
\hline  MAE $L1$  & $0.79$ & $7.39$ & $0.14$ & $0.24 $\\
\hline  MSE $L2$ & $0.76$ & $7.59$ & $0.13$ & $0.25 $\\
\hline Ordinal Reg Loss & $0.90$ & $5.06$ & $0.10$ & $0.15 $\\

\hline
\end{tabular}
\end{center}

\label{change_of_loss}
\end{table}

\begin{figure*}[t]
  \centering
  \includegraphics[width=180mm, height=55mm, scale=0.1]{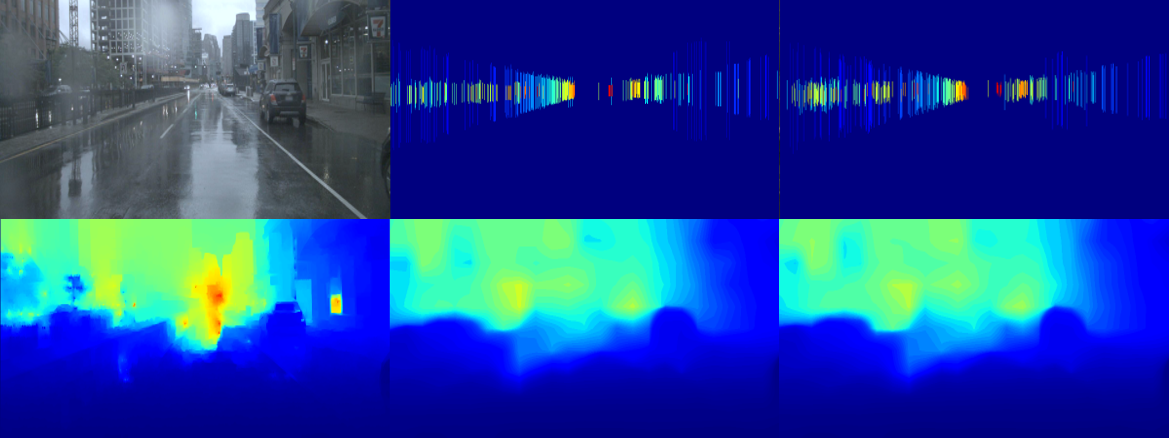}
  \caption{From left to right (a) RGB (b) Height extended radar points with calibration error (c) Calibrated height extended radar points. In second row left to right (a) Ground truth (lidar) (b) Depth Estimation with calibration error of height extended radar points  (c) With calibrated height extended radar points final depth estimation. The qualitative results of depth estimation by actual and interrupted calibration changing the radar camera calibration using RGB+Radar with only extended radar's point cloud.}
  \label{calib-fig}
\end{figure*}
\subsection{Extended Discussion}\label{extended-study}

An essential consideration in the proposed method is sensor calibration, as inaccuracies in calibration parameters can compromise the accuracy of reported results. To evaluate the robustness of our approach, we conducted experiments by intentionally modifying calibration parameters to induce disturbances in sensor transformation, as illustrated in Fig. \ref{calib-fig}. The first row (c) depicts radar markers with synthetically induced noise, while the second row (c) showcases actual radar markers. Remarkably, the depth estimation results (d) demonstrate tolerance to noise induced by calibration errors in sensor parameters. However, it's important to note that the bounds of this tolerance were not extensively explored in this work and could be a focus of future research efforts. Additionally, it's crucial to highlight that the ground truth for depth estimation relies on lidar sensor data. Given that the nuScenes dataset lacks ground truth for monocular camera depth, we rely on lidar sensor information as the ground truth. Therefore, ensuring synchronization between sensors (lidar, camera, and radar) is imperative for achieving more accurate predictions. These findings underscore the importance of rigorous sensor calibration and synchronization for the successful implementation of our proposed method in real-world scenarios.

\section{Conclusion}\label{conclusion}

This study presents the effectiveness of a novel network for radar-validated monocular depth estimation in robotics applications. DepthSense utilizes an innovative approach by leveraging up-sampled lower feature pyramid layers, rather than higher layers, within the network architecture to reduce parameter count while preserving essential features. Through late fusion composition, augmented radar data is seamlessly integrated with RGB image data, providing crucial depth information for addressing the ill-posed estimation problem. The incorporation of an ordinal regression layer transforms the conventional depth regression problem into a classification task, further enhancing accuracy. Our experimental results, validated across various environmental conditions using the complete nuScenes dataset, demonstrate the superior performance of the proposed model compared to \textcolor{black}{SOTA} methods. DepthSense represents a significant advancement over traditional stereo methods, offering a robust and efficient solution for depth estimation in autonomous driving by leveraging the complementary strengths of radar and monocular camera data.


\begin{IEEEbiography}[{\includegraphics[width=1in,height=1.25in,clip,keepaspectratio]{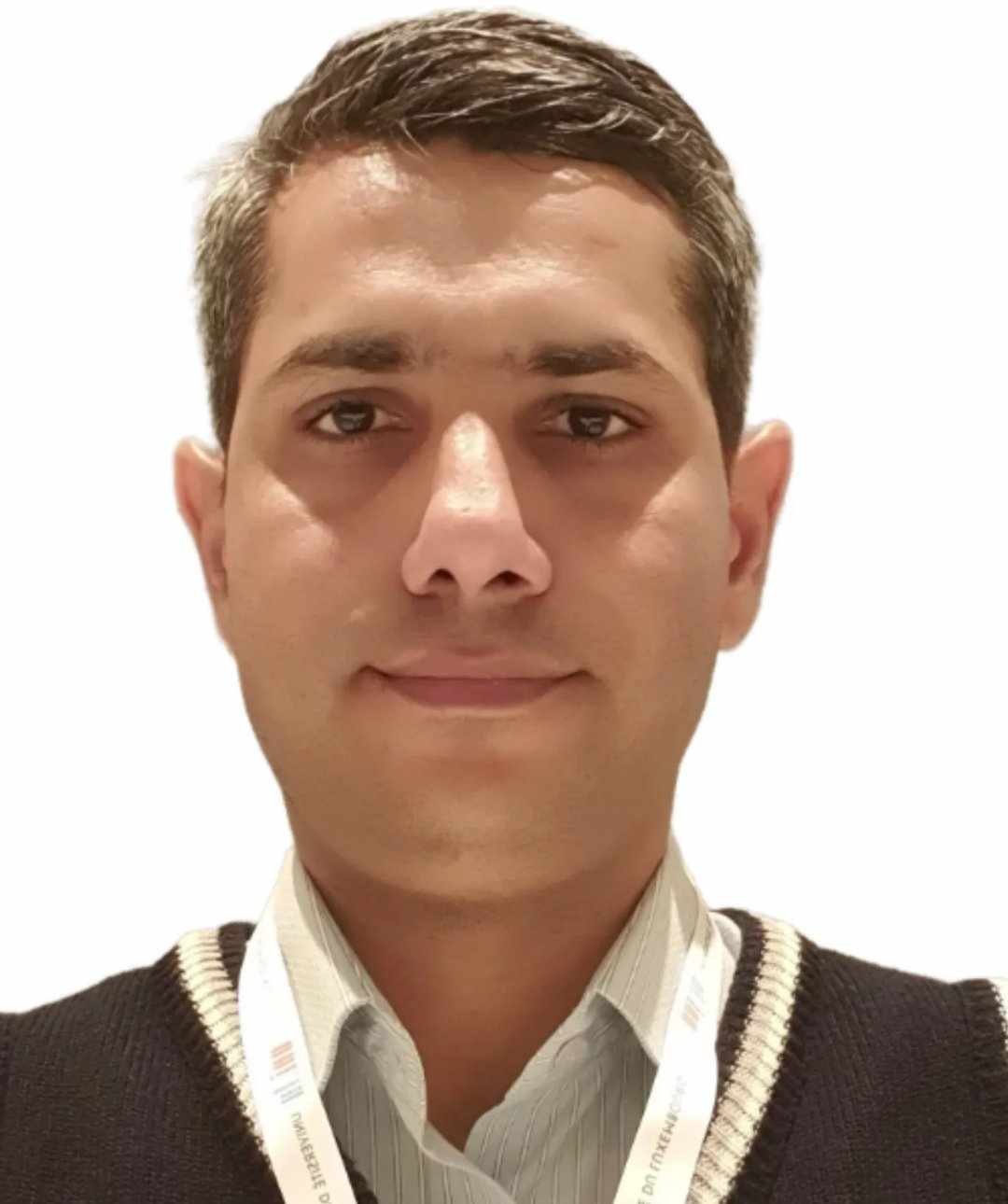}}]{Muhammad Ishfaq Hussain} received his M.S. degree in computer software engineering from the National University of Sciences and Technology, Pakistan, in 2016. In August 2023, he earned a Ph.D. degree from the Machine Learning and Computer Vision lab at the School of Electrical Engineering and Computer Science, Gwangju Institute of Science and Technology, Gwangju, South Korea. Currently, he holds a Post Doc position at the School of Electrical Engineering and Computer Science, Gwangju Institute of Science and Technology. His research interests encompass artificial intelligence, machine learning, sensor fusion, robotics, and autonomous driving.
\end{IEEEbiography}
\begin{IEEEbiography}[{\includegraphics[width=1in,height=1.25in,clip,keepaspectratio]{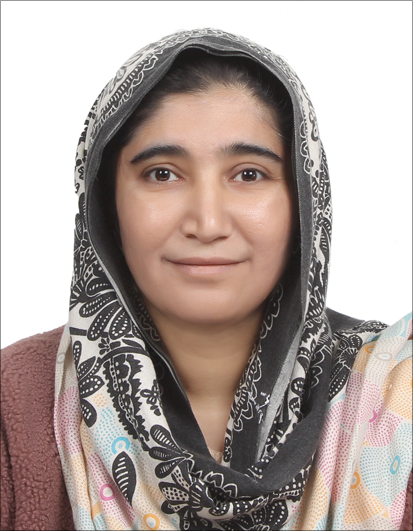}}]{Zubia Naz} received his B.S. degree in information technology from the National University of Sciences and Technology, Pakistan, and from Feb 2023, she is doing M.S degree from the Machine Learning and Computer Vision lab at the School of Electrical Engineering and Computer Science, Gwangju Institute of Science and Technology, Gwangju, South Korea. Her research interests encompass artificial intelligence and machine learning.
\end{IEEEbiography}

\begin{IEEEbiography}[{\includegraphics[width=1in,height=1.25in,clip,keepaspectratio]{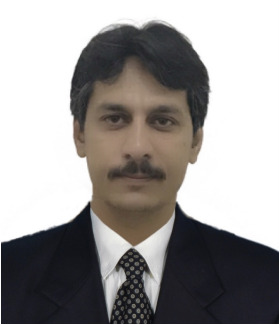}}]{Muhamamd Aasim Rafique} received his M.Sc. degree in computer science from Quaid-e-Azam University, Islamabad, Pakistan. He then received his M.Sc. degree in computer science from Lahore University of Management and Sciences, Lahore, Pakistan, in 2008. He received his Ph.D. degree from the School of Electrical Engineering and Computer Sciences, GIST, Gwangju, Republic of Korea, in 2018. He is an Assistant Professor in the School of Electrical Engineering and Computer Science, National University of Sciences and Technology
(NUST), Pakistan and currently onleave to attend PostDoc at GIST. His research interests lie in artificial neural networks and their applications in machine learning and computer vision.
\end{IEEEbiography}

\begin{IEEEbiography}[{\includegraphics[width=1in,height=1.25in,clip,keepaspectratio]{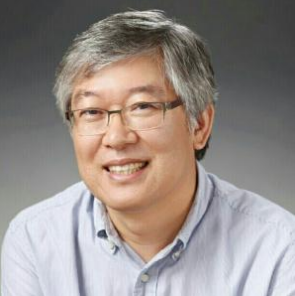}}]{Moongu Jeon} received a B.S. degree in architectural engineering from Korea University, Seoul,
South Korea, in 1988 and M.S. and Ph.D. degrees in computer science and scientific computation from the University of Minnesota, Minneapolis, MN, USA, in 1999 and 2001, respectively. As a postgraduate researcher, he worked on optimal control problems at the University of California at Santa Barbara, Santa Barbara, CA, USA, from 2001 to 2003 and then moved to the National Research Council of Canada, where he worked on the sparse representation of high-dimensional data and image processing until 2005. In 2005, he joined the Gwangju Institute of Science and Technology, Gwangju, South Korea, where he is currently a Full Professor with the School of Electrical Engineering and Computer Science. His current research interests lie in machine learning, computer vision, and artificial intelligence.
\end{IEEEbiography}

\end{document}